\documentclass[sigconf]{acmart}
\usepackage{diagbox}
\renewcommand\footnotetextcopyrightpermission[1]{}
\usepackage{algorithm}
\usepackage{algorithmic}
\usepackage{amsmath}

\begin{document}

\title{Cued-Agent: A Collaborative Multi-Agent System for Automatic Cued Speech Recognition}



\author{Guanjie Huang}
\orcid{0000-0003-1977-3647}
\affiliation{%
  \institution{The Hong Kong University of Science and Technology (Guangzhou)}
  \city{Guangzhou}
  \country{China}
  }
\email{ghuang565@connect.hkust-gz.edu.cn}

\author{Danny H.K.~Tsang}
\orcid{0000-0003-0135-7098}
\affiliation{%
  \institution{The Hong Kong University of Science and Technology (Guangzhou)}
  \city{Guangzhou}
  \country{China}
  }

\email{eetsang@ust.hk}

\author{Shan Yang}
\orcid{0000-0003-4464-146X}
\affiliation{
    \institution{Tencent AI Lab}
    \city{Shenzhen}
    \country{China}
}
\email{shaanyang@tencent.com}

\author{Guangzhi Lei}
\orcid{0009-0009-6136-7619}
\affiliation{
    \institution{Tencent AI Lab}
    \city{Shenzhen}
    \country{China}  
}
\email{guangzhilei@tencent.com}

\author{Li Liu}
\orcid{0000-0002-4497-0135}
\authornote{Corresponding Author}
\affiliation{%
  \institution{The Hong Kong University of Science and Technology (Guangzhou)}
  \city{Guangzhou}
  \country{China}
  }
\email{avrillliu@hkust-gz.edu.cn}





\begin{abstract}
    Cued Speech (CS) is a visual communication system that combines lip-reading with hand coding to facilitate communication for individuals with hearing impairments. Automatic CS Recognition (ACSR) aims to convert CS hand gestures and lip movements into text via AI-driven methods. Traditionally, the temporal asynchrony between hand and lip movements requires the design of complex modules to facilitate effective multimodal fusion. However, constrained by limited data availability, current methods demonstrate insufficient capacity for adequately training these fusion mechanisms, resulting in suboptimal performance. Recently, multi-agent systems have shown promising capabilities in handling complex tasks with limited data availability. To this end, 
    we propose the first collaborative multi-agent system for ACSR, named \textbf{Cued-Agent}. It integrates four specialized sub-agents: a Multimodal Large Language Model-based Hand Recognition agent that employs keyframe screening and CS expert prompt strategies to decode hand movements, a pretrained Transformer-based Lip Recognition agent that extracts lip features from the input video, a Hand Prompt Decoding agent that dynamically integrates hand prompts with lip features during inference in a training-free manner, and a Self-Correction Phoneme-to-Word agent that enables post-process and end-to-end conversion from phoneme sequences to natural language sentences for the first time through semantic refinement. To support this study, we expand the existing Mandarin CS dataset by collecting data from eight hearing-impaired cuers\footnote{Cuer represents the person who performs CS.}, establishing a mixed dataset of fourteen subjects. Extensive experiments demonstrate that our Cued-Agent performs superbly in both normal and hearing-impaired scenarios compared with state-of-the-art methods. The implementation is available at \href{https://github.com/DennisHgj/Cued-Agent}{\textit{https://github.com/DennisHgj/Cued-Agent}}.

\end{abstract}

\maketitle

\section{Introduction}
\label{sec: Introduction}
\begin{figure}[t]
\includegraphics[width=1.0\linewidth]{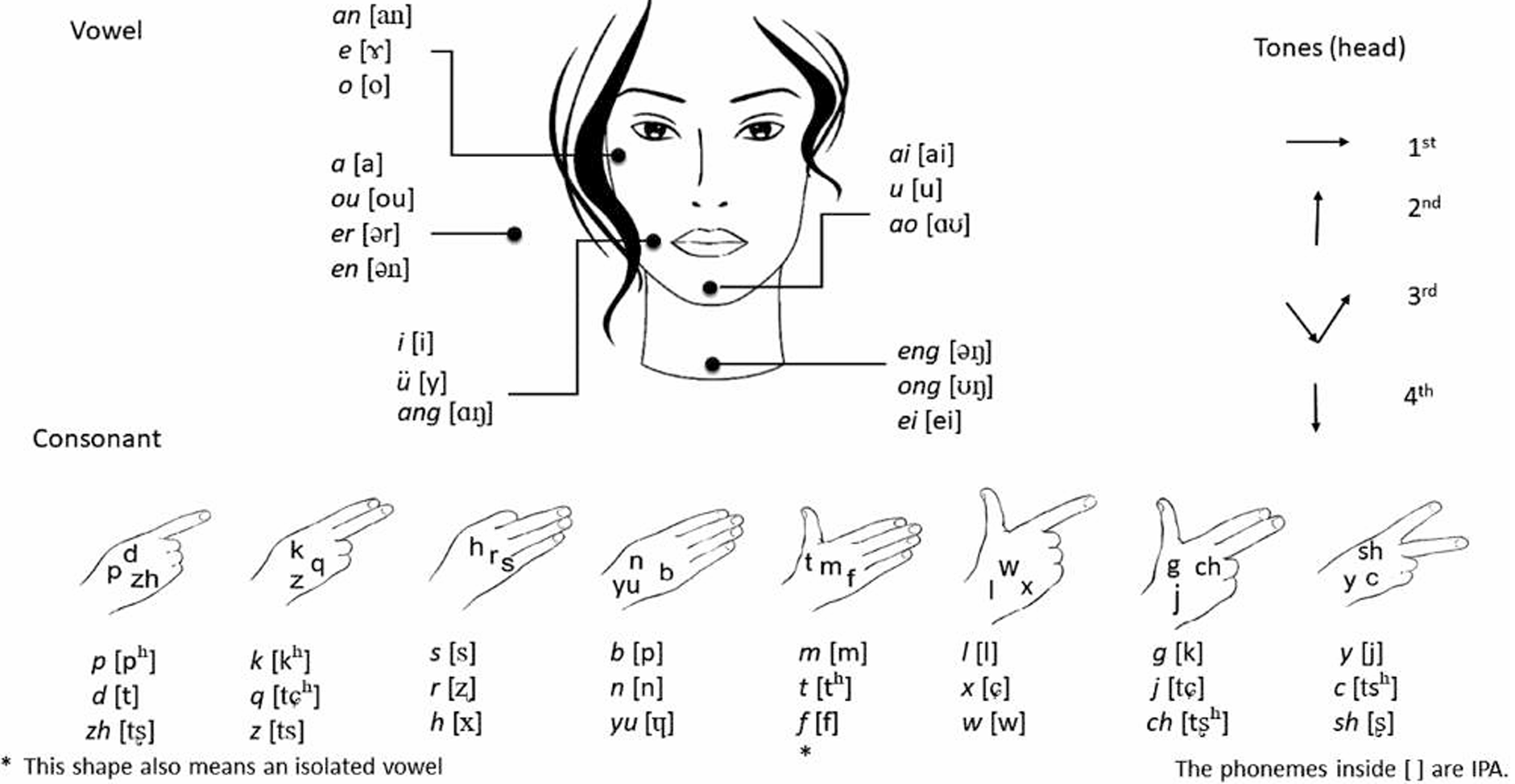}
\centering
\caption{Mandarin Chinese CS system. Hand coding contains five hand positions and eight hand shapes to encode Mandarin vowels and consonants for assisting lip reading (image from \cite{liu2019pilot}).}
\label{fig:mandarinCS} 
\end{figure}

\begin{figure*}[ht]
\includegraphics[width=1.0\linewidth]{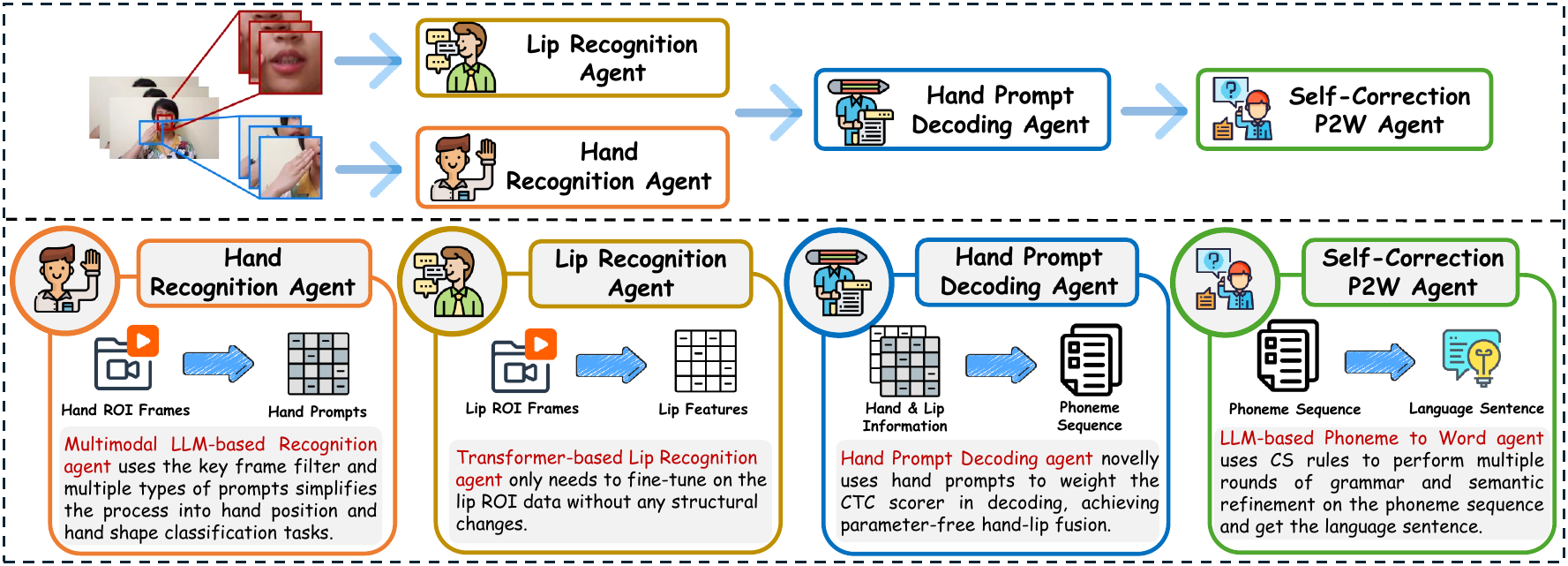}
\centering
\caption{Overview of the proposed Cued-Agent. Four agents (i.e., Hand Recognition Agent, Lip Recognition Agent, Hand Prompt Decoding Agent, and Self-Correction P2W Agent) cooperate for ACSR. 
}
\label{fig:framework} 
\end{figure*}

Lip-reading poses significant challenges for hearing-impaired individuals due to the inherent ambiguity of visual speech information. To address this limitation, the first Cued Speech (CS) system was introduced in 1967 for American English \cite{cued1967}, augmenting lip-reading with hand gestures to represent phonetic elements visually. Since its inception, CS has proven highly adaptable and has been customized for more than 65 languages worldwide~\cite{trezek2017cued,ling1975cued,cornett1994adapting}. Notably, \cite{liu2019pilot} pioneered the Mandarin Chinese CS system (Fig.~\ref{fig:mandarinCS}), which employs five hand positions to encode Chinese vowel groups and eight hand shapes to distinguish 24 consonant phonemes. This innovation has empowered Mandarin-speaking hearing-impaired individuals by enhancing their access to spoken language through combined visual and gestural cues.

With advancements in deep learning, Automatic CS Recognition (ACSR)~\cite{HERACLEOUS2010504, liu2018automatic} has emerged as an essential research area, owing to its great potential to aid daily communication for the hearing-impaired. ACSR aims to convert CS video inputs, comprising lip movements and hand gestures, into text output. Common approaches leverage Transformer-based architectures to develop cross-modal fusion strategies \cite{liu2023cross,liu2024computation}, which seek to harmonize complementary relationships between lip and hand signals for accurate cued phoneme sequence prediction. However, the performance of such methods remains constrained by the limited scale of existing CS datasets, which hinders the training of sophisticated fusion modules. 
Recent work ~\cite{huang2025lendhandsemitrainingfree} abandons complex Transformer-based fusion modules and leverages a Multimodal Large Language Model (MLLM) to obtain hand information. However, thousands of hand recognition results from the training set still need to be obtained from MLLM to train the hand-lip fusion layer, which is costly.

Despite this, all prior ACSR methods have been restricted to output phoneme sequences rather than natural language sentences. Although phonemes encode the accurate pronunciation, they lack natural language's semantic and structural richness, resulting in ambiguous or incomplete interpretations. Existing Phoneme-to-Word (P2W) tools can achieve the conversion from phoneme sequences to word sentences~\cite{hsu1999phoneme}. However, their lack of self-correction capabilities makes it difficult to adapt to ACSR tasks. Predicted phoneme sequences with slight errors cause significant difficulties in the conversion. This mismatch creates recognition barriers that phoneme-based outputs struggle to align with real-world sentences, making interactions non-intuitive and error-prone.

Recently, auto-agent-based paradigms have shown potential in handling complex tasks and data-scarce scenarios~\cite{li2024anim,tu2024spagent}. To retain the CS system's advantage in communication accuracy and address the above limitations, in this work, we fully exploit the capabilities of LLM-based agents and build a multi-agent system suitable for ACSR tasks, named Cued-Agent. This is the first work to achieve parameter-free fusion of hand and lip modalities and automatically correct the phoneme sequence to obtain the corresponding sentence. To be more specific, Cued-Agent contains four specialized agents. \textbf{Firstly}, an MLLM-based Hand Recognition Agent takes hand region of interest (ROI) frames as input and outputs \textcolor{black}{hand prompts, which are a phoneme encoding matrix containing the position and shape recognition results.}. It designs a keyframe filter and multiple specialized prompts to simplify the recognition into hand position and shape classification tasks. \textbf{Secondly}, a Transformer-based Lip Recognition agent extracts high-quality lip features from lip ROI frames. \textbf{Thirdly}, a Hand Prompt Decoding agent takes hand prompts and lip features as input and outputs the phoneme sequence. It equips hand prompts to dynamically weight the decoding score in the beam search process, thus achieving parameter-free hand-lip fusion without any further training. \textbf{Lastly}, a Large Language Model (LLM)-based Self-Correction P2W agent takes the phoneme sequence as input, then outputs the revised sequence and corresponding language sentence. The agent automatically performs multiple grammar and semantic refinement rounds based on specialized CS rule prompts.

To summarize, the main contributions are as follows: 
\begin{itemize}
    \item We propose a new multi-agent ACSR system (Cued-Agent) for the first time. With the cooperation of four specialized sub-agents, Cued-Agent achieves training-free hand information recognition, parameter-free hand-lip fusion, and self-correcting conversion from CS videos to language sentences.
    
    \item Instead of costly training a hand-lip fusion module, the proposed Hand Prompt Decoding agent uses hand information to weight the decoding process, realizing the fusion of hand-lip information without training parameters.

    \item We propose the Self-Correction P2W agent, which realizes the self-correction conversion from phoneme sequences to natural sentences. We also bring two new indicators to comprehensively evaluate the conversion results from the perspectives of word accuracy and semantic similarity.

    \item To support this study, we especially expanded data collection by incorporating recordings from eight hearing-impaired Mandarin CS cuers to form the largest hearing-impaired CS dataset. It provides the most realistic evaluation of hearing-impaired scenes. Through comprehensive experiments, our proposed Cued-Agent demonstrates superior performance across diverse experimental configurations.
\end{itemize}




\section{Related Works}
\begin{figure}[t]
\includegraphics[width=0.95\linewidth]{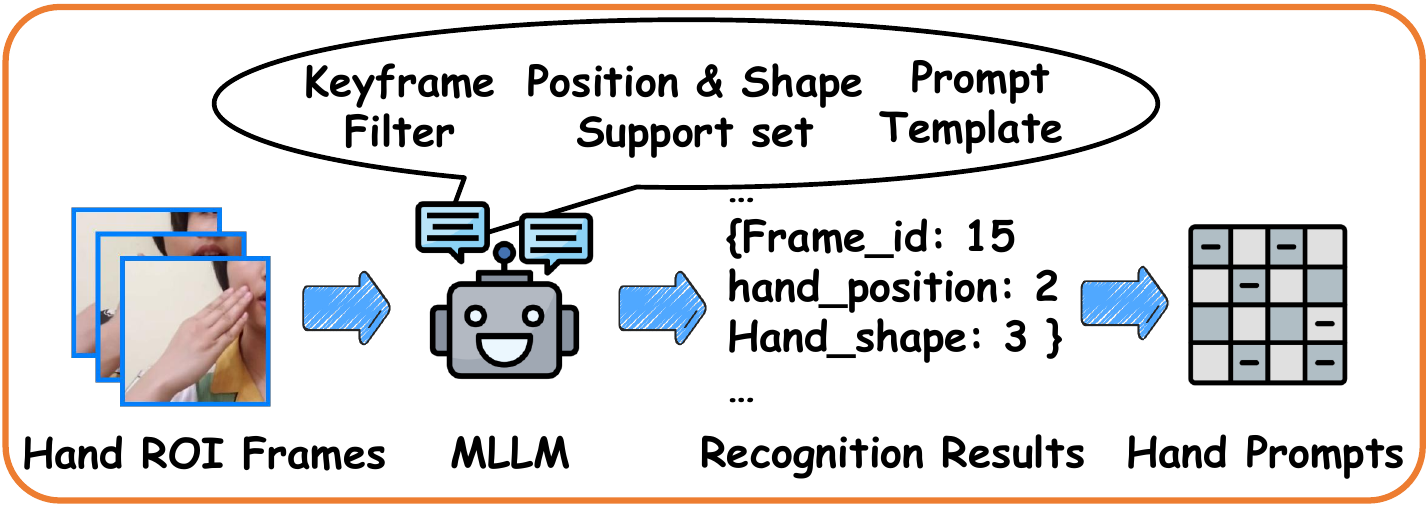}
\centering
\caption{Hand Recognition Agent.}
\label{fig:hand_agent}
\end{figure}

\label{sec:Related Works}
\subsection{Automatic Cued Speech Recognition} ACSR has evolved significantly since its inception~\cite{heracleous2010cued}. In 2018, \cite{liu2018automatic} pioneered the application of Deep Learning models to ACSR, establishing a foundation for subsequent research. Traditional approaches in ACSR emphasize designing fusion modules to address the inherent asynchrony between lip and hand modalities, which arise from the temporal misalignment of visual and gestural cues in CS videos. Early studies, such as \cite{heracleous2012continuous}, \cite{9413407}, \cite{liu2018visual}, \cite{9287365}, \cite{4802159} and \cite{wang2021attention}, focused on extracting ROI features from lip and hand modalities and concatenating them directly for cross-modal fusion, but resulted in a lack of robustness to temporal discrepancies. To mitigate this, \cite{gao2023novel} introduced a re-synchronization process that predicts hand-preceding time intervals in CS datasets. 

Recent Transformers~\cite{vaswani2017attention} have excelled in capturing global dependencies, making them an ideal choice for solving the problem of asynchronous multimodal fusion in ACSR.
For instance, \cite{liu2023cross} proposed a hand-lip mutual learning framework using cross-attention mechanisms. This approach resolves modality asynchrony and eliminates the need for phoneme-level annotations, relying instead on sentence-level supervision. However, the high computational and parameter costs of Transformers posed challenges. Addressing this, \cite{liu2024computation} optimized the Transformer architecture by refining the fusion module design, achieving good performance while reducing computational complexity. These works typically employ pre-trained CNN front-ends for feature extraction and custom Transformer-based fusion modules trained from scratch. 

Most recent work~\cite{huang2025lendhandsemitrainingfree} abandoned the Transformer-based fusion module and hand modeling. Instead, it 
designed a special module to extract keyframes from the hand ROI frames, and combined designed prompts with the generalization ability of MLLM to simplify the hand recognition task into a training-free shape and position classification task. 
However, the cost of training the hand-lip fusion layer is still high, which requires obtaining all hand recognition results of the entire training set through extensive MLLM inferences.

Besides, all the above works limit the model to outputting phoneme sequences rather than actual sentences, which leads to communication barriers and application difficulties.

\subsection{Multi-agent System}

A multi-agent system is a distributed system composed of multiple autonomous or semi-autonomous agents that work together to complete complex tasks through collaboration, communication, and coordination~\cite{han2024llm,van2008multi}. In recent years, with the continuous development of LLMs, many works have used LLM-based agents as the core AI agents to form a multi-agent system~\cite{li2024survey,guo2024large,he2024llm}. The idea of using multiple agents to solve complex tasks has also received widespread attention and has been applied to multiple subtasks. These include image generation and editing~\cite{wang2024genartist,li2025mccd,10.1145/3731715.3733271}, video generation~\cite{li2024anim,tu2024spagent}, video understanding~\cite{fan2024videoagent, wang2021attention} and audio generation~\cite{huang2024audiogpt,liang2024wavcraft,zhang2025long}.

However, this design has not been fully explored in ACSR. The CS system uses hand encoding to assist lip-reading. These two asynchronous modalities are naturally suitable for processing using different agents. Therefore, Cued-Agent first utilized the concept of multi-agent systems to divide the originally complex ACSR task into multiple subtasks, and employed specialized agents to handle them. The new parameter-free decoding method solves the shortcoming that the fusion module needs joint training and increases the flexibility of the overall system.

\section{Methods}
\label{sec:method}
\begin{algorithm}[t!]
\caption{Grouping process of the keyframe filter in Hand Recognition agent.\\
\textbf{Input}: Slow-motion frame set $P$\\
\textbf{Parameter}:  Index distance threshold $\theta$
} 
\label{alg:group}
\begin{algorithmic}[1]
    \STATE Let $k=0, m=0$
    \STATE Initialize current slow-motion group list $G_k$
    \FOR{$X^h_j$ in $P$}
    \IF {$G_k =\emptyset $}
        \STATE $G_k \leftarrow G_k \cup {X^h_j}, m=j$
    \ELSIF {$j-m \leq \theta$}
        \STATE $G_k \leftarrow G_k \cup {X^h_j}, m=j $
    \ELSE 
        \STATE $G \leftarrow G \cup {G_k}, k =k+1$ 
    \ENDIF
    \ENDFOR
    \STATE \textbf{return} $G$
    \end{algorithmic}
\end{algorithm}

\subsection{Problem Formulation}
The CS dataset contains $N$ videos and corresponding sentence-level annotations, denoted by $Z = \{[X_i,  Y^p_i,  Y^a_i]\}^N_{i=1}$, where $Y^p_i$ is the phoneme sequence label and $Y^a_i$ is the word sentence. After preprocessing, CS videos are transferred to lip ROI frames, hand ROI frames, and hand position sequences, which are denoted by $Z = \{[X_i^{l}, X_i^{h}, X_i^{r},  Y^p_i,  Y^a_i]\}^N_{i=1}$. Specifically, for a $T$ frame video, lip ROI video $X_i^{l} \in \mathbb{R}^{T\times W^l \times H^l}$ and hand ROI video $X_i^{h} \in \mathbb{R}^{T\times W^h \times H^h}$, where $W^l, H^l, W^h, H^h$ are the width and height of lip and hand ROI frames. Hand position sequence $X_i^{r} \in \mathbb{R}^{T\times 2}$ indicates the center point coordinates of the hand in video frames. ACSR's target is to form a system that maps the multimodal data input $(X^{l}, X^{h}, X^{r})$ to the corresponding phoneme sequence and sentence $Y^p_i,  Y^a_i$.

\subsection{Overview of Cued-Agent}
Cued-Agent is a multi-agent system that leverages four different agents to achieve recognition from visual inputs to language sentences. Illustrated in Fig.~\ref{fig:framework}, the original CS video is preprocessed into lip and hand ROI frames. After that, the Hand Recognition and Lip Recognition agents process the ROI videos separately and output information to the Hand Prompt Decoding agent to perform fusion decoding and obtain the phoneme sequence. The Self-Correction P2W agent combines the CS system rules and semantic fluency to perform multiple rounds of modification on the phoneme sequence, and finally obtains the corrected phoneme sequence and the corresponding natural language sentence.
\begin{figure}[t]
\includegraphics[width=0.75\linewidth]{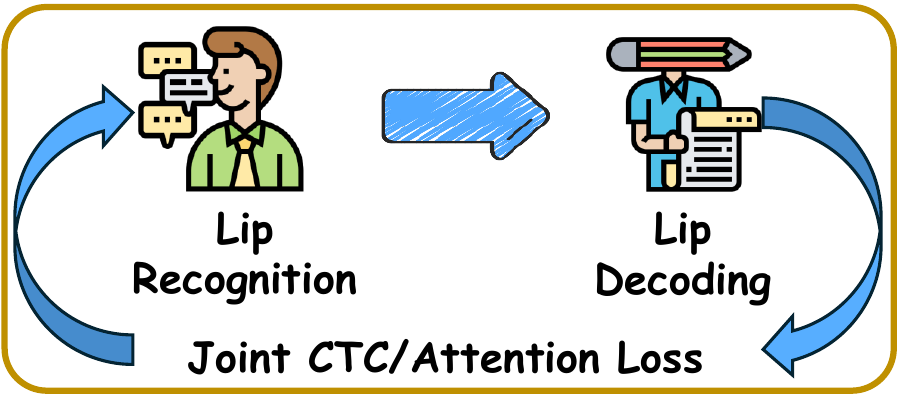}
\centering
\caption{Lip Recognition Finetuning Process.}
\label{fig:lip_agent}
\end{figure}

\subsection{Hand Recognition Agent}

As shown in Fig.~\ref{fig:hand_agent}, the Hand Recognition agent inputs the hand ROI frames and designs a special prompt module to help MLLM recognize the hand position and shapes in keyframes. Then, an embedding function encodes the recognition results into a hand prompt matrix as the final output. Following ~\cite{huang2025lendhandsemitrainingfree}, the prompt module effectively simplifies the hand movement recognition task into the keyframe position and shape classification tasks through the following components: a hand keyframe filter, a hand position and shape support set, and a customized prompt template.

\noindent \textbf{Hand Keyframe Filter.} The CS system encodes different phonemes using limited hand positions and shapes, and the user will pause at the key phoneme position for a while. Therefore, the filter selects slow-motion frames as keyframes for hand recognition based on the speed of hand movement. Specifically, when a hand ROI video $X^{h}$ and the corresponding hand position sequence $X^{r}$ are input, the filter will screen out keyframes through the following steps:

\textit{Slow-motion selection.} The filter evaluates hand-moving speed by calculating the movement distance $D_j$ between adjacent frames:
\begin{equation}
 D_j = distance(X_{j-1}^{r},X_j^{r}), j\in[1,T),
\label{equa:distance}
\end{equation}
where function $distance(\cdot, \cdot)$ here is the Euclidean distance. The slow-motion frame set $P$ is filtered out by the moving distance and threshold $\sigma$:
\begin{equation}
 X_j^{h} \in P \mid D_j \leq \sigma.
\end{equation}

\textit{Grouping and deduplication.} To avoid recognizing adjacent slow-motion frames repeatedly, the filter performs grouping and deduplication operations on all slow-motion frames and selects one frame as the keyframe of each group. The filter follows Alg.~\ref{alg:group} to group slow-motion frames and get the slow-motion group set $G$ that contains $M$ groups. In simple terms, the difference between the frame index numbers of adjacent frames in the same group cannot be greater than a threshold $\theta$. For each slow-motion group, select the middle frame as the keyframe and put it into the keyframe set $K$.

\noindent \textbf{Hand Position and Shape Support Set.}
To fully leverage the in-context learning capability of MLLM, a support set that contains all combinations of hand position and shape is equipped to provide a direct visual reference. 

\noindent \textbf{Customized Prompt Template.} In each inference process, the Hand Recognition agent puts the keyframe set $K$ and the support set into a customized prompt template to obtain recognition results from MLLM. The prompt templates can be divided into four types of prompts according to their functions. 

\textit{Background Prompts.} These prompts help MLLM grasp the specific role and context it needs to adopt. They state MLLM should be a specialist in detecting hand positions and shapes within keyframes of CS videos. The prompt also outlines how the CS system works and why hand movement analysis is crucial.

\textit{Multimodal In-context Prompts.} In addition to a detailed textual definition of the position and shape of the hand, prompts also use images from the support set to construct reference images for each category, attempting to establish a correspondence between data and labels in a multimodal way.

\textit{Contrastive Prompts.} Since both hand positions and shapes have easily confused categories, these prompts improve MLLM's recognition accuracy by contrasting these categories through key distinguishing features and combining visual examples to demonstrate differentiation methods.

\textit{Chain-of-Thought Prompts.} They use step-by-step reasoning to guide MLLM through hand analysis. For keyframes in set $K$, prompts first require the MLLM to compare hand position and shape with the support set using clear logical steps, then double-check the results by matching them with text descriptions of each category.

\noindent \textbf{Embedding Function.} After getting recognition results from MLLM for all $M$ keyframes of set $K$, the embedding function will transfer the hand and position category to phonemes according to the CS hand encoding rules. Then, a matrix $H \in \mathbb{R}^{T \times q}$ initialized to zeros is used to encode the hand information in the one-hot coding manner. The dimension $T$ is the original frame number, and $q$ is the number of phonemes and necessary symbols. For each keyframe $X^h_m$ and frames in the same slow-motion group $G_m$, the matrix will set the phoneme value in the vector corresponding to each frame to one according to the keyframe recognition result. After encoding all hand recognition results, the matrix $H \in \mathbb{R}^{T \times q}$ will be output as the hand prompt for the decoding agent.

\begin{figure}[t]
\includegraphics[width=0.9\linewidth]{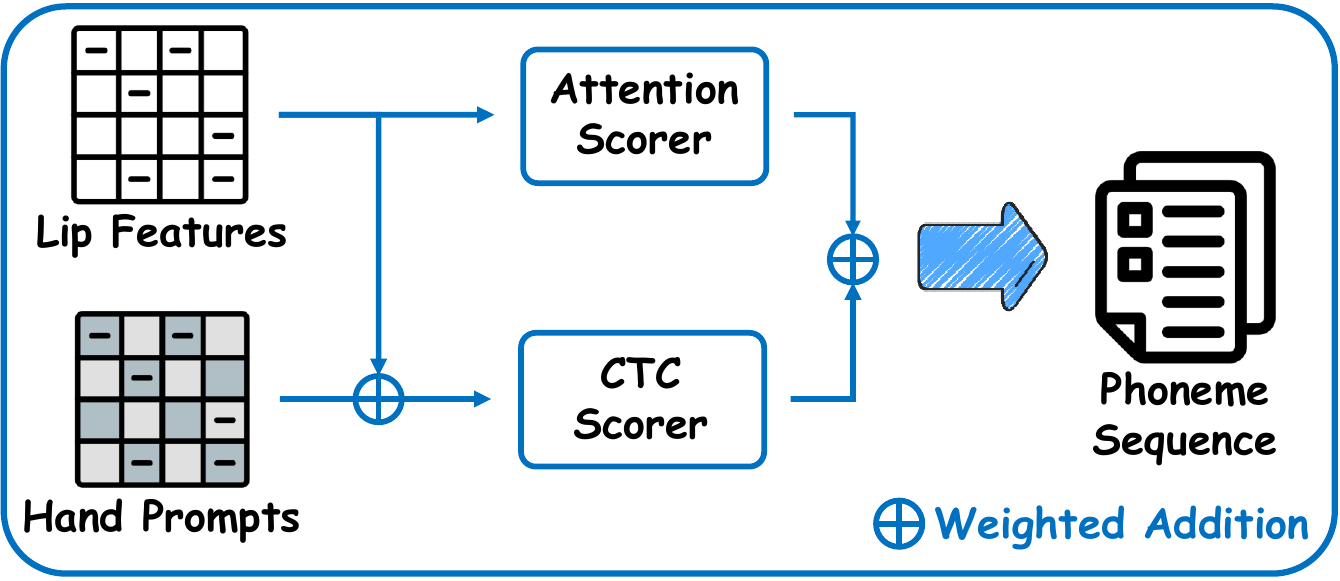}
\centering
\caption{Parameter-Free Hand-Lip Joint Decoding in the Hand Prompt Decoding Agent.}
\label{fig:decoding_agent} 
\end{figure}

\begin{figure*}[t]
\includegraphics[width=0.9\linewidth]{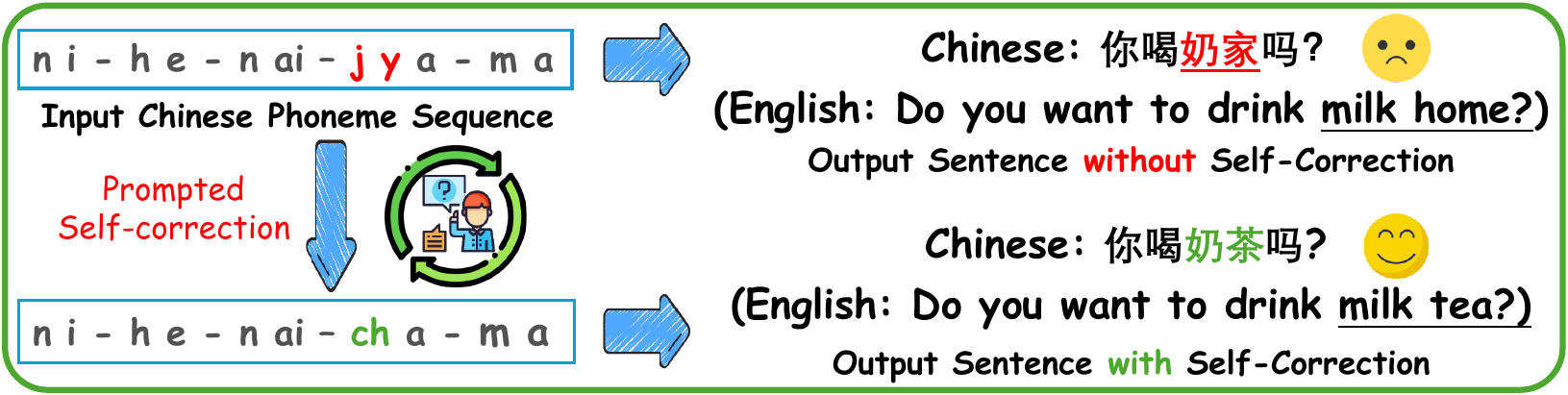}
\centering
\caption{Self-Correction Phoneme-to-Word Agent. It shows a real self-correction example: with the help of customized prompts, the agent makes minor corrections to the phoneme sequence and outputs a semantically coherent Mandarin sentence.}
\label{fig:P2W_agent} 
\end{figure*}

\subsection{Lip Recognition Agent}

Transformer-based Lip Recognition agent takes lip ROI frames $X^{l}$ as input and outputs the lip features $L \in \mathbb{R}^{T \times d}$, where $d$ is the feature dimension. To obtain better lip-reading recognition results, the pre-trained Lip Recognition agent needs to be finetuned with the decoder under the visual speech recognition paradigm using only lip ROI frames $X^{l}$ and the corresponding phoneme sequence labels $Y^p$. As shown in Fig.~\ref{fig:lip_agent}, we use joint Connectionist Temporal Classification (CTC)/attention training~\cite{watanabe2017hybrid} to finetune the two agents. The loss is calculated by:
\begin{equation}
    \mathcal{L}_{\mathrm{lip}}=\lambda_{train} \log p_{\mathrm{ctc}}(Y^p\mid X^l)+(1-\lambda_{train}) \log p_{\mathrm{att}}(Y^p\mid X^l),
\end{equation}
where the tunable parameter $\lambda_{train}$ satisfies $0 \leq \lambda_{train} \leq 1 $, and $p_{\mathrm{ctc}}$ and $p_{\mathrm{att}}$ are the sequence posteriors for the CTC task and the attention decoder.

\subsection{Hand Prompt Decoding Agent}
During the inference, the finetuned decoder leverages the hand prompts to weight the decoding procedure, thus achieving parameter-free hand-lip fusion decoding. We named it the Hand Prompt Decoding agent.
As shown in Fig.~\ref{fig:decoding_agent}, the agent takes hand prompts $H$ and lip feature $L$ as input to decode the phoneme sequence $S^{p'}$.  
Specifically, in the beam search process, the CTC score is defined as 
\begin{equation}
    \alpha_{\mathrm{ctc}}(h, L)  \triangleq \log p_{\mathrm{ctc}}(h, ... \mid L),
\end{equation}
where $h$ is the hypothesis and the $\log p_{\mathrm{att}}(h, ... \mid L)$ 
is the CTC prefix probability~\cite{graves2012supervised} defined as the cumulative probability of all label sequences that have $h$ as their prefix:
\begin{equation}
    p_{\mathrm{ctc}}(h, \ldots \mid L)=\sum_{\nu \in(\mathcal{Q} \cup\{<\mathrm{eos}>\})} p_{\mathrm{ctc}}(h \cdot \nu \mid L),
\end{equation}
where $\nu$ represents possible character options, and $Q$ represents the set of all phonemes and symbols except the <eos>, with a total size of $q-1$. $p_{\mathrm{ctc}}$ is the CTC posterior. Therefore, to comply with the rules for CTC score calculation and avoid matrix operations with inconsistent dimensions, $L$ here actually should be the $L' \in \mathbb{R}^{T \times q}$, which maps by a trained linear layer from the original $L \in \mathbb{R}^{T \times d}$. 

Here, we innovatively add hand prompts $H$ and $L'$ with the same dimension to obtain a new hand-lip fusion feature $L^H$ without any parameters or further training. It is denoted by:
\begin{equation}
   L^H = L' + \lambda_{prompt}*H, 
\end{equation} 
where $\lambda_{prompt}\geq 0$. The CTC score here is redefined as:
\begin{equation}
    \alpha_{\mathrm{ctc}}(h, L^H)  \triangleq \log p_{\mathrm{ctc}}(h, ... \mid L^H),
\end{equation}
and the joint scoring process in beam search to obtain the final phoneme sequence $ S^{p'}$  is defined as:
\begin{equation}
\begin{split}
 S^{p'} = \arg \max _{h \in\hat{\Omega}} \biggl\{\lambda_{\text{decode}} \alpha_{\text{ctc}}(h, L^H) + (1 - \lambda_{\text{decode}}) \alpha_{\text{att}}(h, L) \biggr\},
\end{split}
\end{equation} 
where $\hat{\Omega}$ denotes a set of complete hypotheses, the tunable parameter $\lambda_{decode}$, satisfies $0 \leq \lambda_{decode} \leq 1$, and $\alpha_{\text{att}}(h \mid L) $ is the attention score which is defined as:
\begin{equation}
    \alpha_{\mathrm{att}}(h, L)  \triangleq \log p_{\mathrm{att}}(h \mid L).
\end{equation}

\begin{table*}[t] 
\caption{Details of different language-based CS Datasets. The ``Cuer'' row shows the subset with different cuer settings in datasets. The number indicates the cuer number, ``HI'' or``H'' means cuers are hearing-impaired or not. Our newly proposed MHI-MCCSD is the first large-scale multi-hearing-impaired cuer dataset.}
\centering

\tabcolsep 0.17in
\begin{tabular}{c|c|cc|ccc|c}
\toprule[1.3pt]
Dataset & French & \multicolumn{2}{|c|}{British} & \multicolumn{3}{|c|}{MCCSD} & \textbf{MHI-MCCSD}\\
\midrule[1pt]
Cuer & 1-H & 1-H & 5-H & 1-H & 1-HI & 6-H  & \textbf{8-HI}\\
\midrule[1pt]
Sentence & 238 & 97 & 390 & 1,000 & 818 & 6,000 & 5,272\\
Character &12,872&2,741&11,021  &32,902 & 8,269 & 197,412 & 152,921\\
Word & -&- &- & 10,564 & 8,269 &63,384 &49,280\\
Phoneme &35&44&44&40&40&40&40 \\
Train&193&78&312&800&652&4,800&4,220\\
Test& 45&19&78&200&166&1,200&1,052\\
\bottomrule[1.3pt]
\end{tabular}
\label{table:dataset}
\end{table*}

\subsection{Self-Correction Phoneme-to-Word Agent}
After getting the phoneme sequence $S^{p'}$ from the decoding agent, the LLM-based Self-Correction P2W Agent performs multiple rounds of self-correction on the phoneme sequence, thus obtaining the revised phoneme sequence $S^{p}$ and the corresponding natural sentence $S^{a}$. Fig.~\ref{fig:P2W_agent} shows a real self-correction example. With the help of customized prompts, the agent combined CS rules and language capabilities to make minor corrections to the phoneme sequence and finally output a semantically coherent sentence. According to their functions, prompts can be divided into the following categories:

\textit{Background Prompts.} These prompts provide an overview of the CS system, explaining its purpose and background. They emphasize that the key responsibility of the LLM is to refine potentially inaccurate phoneme sequences through post-processing, aiming to transform them into semantically coherent sentences. 

\textit{Conversion Rules.} These prompts mainly explain the rules for converting phonemes to text. In the CS system, the encoding of phonemes as text pronunciations may not be entirely consistent with the commonly used pronunciation and spelling. Especially in Chinese, phoneme combinations are not entirely equivalent to pinyin sequences in some scenarios. This part of the prompt details inconsistencies and explains the interchange rules.

\textit{In-context and Contrastive Prompts.} The prompts contain dozens of phoneme sequences and their corresponding sentences in the training set to stimulate LLM's in-context learning ability and help it understand the phoneme sequences' format and conversion relationship. Similarly, the prompts also contain several easily confused phoneme pairs. They often correspond to the same hand encoding, so LLM should first consider the correction of easily confused pairs during the correction process.

\textit{Task Prompts.} The prompts again clarify the task. First, LLM should check and correct the obvious grammar errors within the phoneme sequence. Second, try to translate it into a sentence. If difficulties are encountered or the final sentence is not semantically coherent, try to make minimal modifications to improve it.


\subsection{New Sentence Evaluation Indicators}
As Cued-Agent first realizes the recognition from CS videos to natural sentences, we proposed two new evaluation indicators that can directly evaluate the quality of natural sentences, namely \textbf{Sentence-Word Error Rate (S-WER)} and \textbf{Semantic Score}, respectively.

The S-WER directly measures the word error rate of predicted sentences. It follows the calculation way of Character Error Rate (CER) and Word Error Rate (WER), which are commonly used indicators for phoneme sequence quality evaluation, but performs word segmentation on the sentence.

Semantic Score reports the semantic similarity of predicted sentences and the ground truth. It is denoted as:
\begin{equation}
    Semantic\ Score = \frac{1}{N}\sum_{i=1}^{N} \frac{\mathcal{M}(S^{a}_i)\cdot\mathcal{M}( Y^a_i)}{||\mathcal{M}(S^{a}_i)||||\mathcal{M}( Y^a_i)||},
\end{equation}
where $\mathcal{M}(\cdot)$ is a pretrained language embedding model. Semantic Score can evaluate whether the output sentences are closer to the semantic information based on semantic similarity.

\section{Multi-Hearing-Impaired Mandarin Chinese CS Dataset}

Current CS datasets include the Mandarin Chinese CS Dataset (MCCSD)~\cite{liu2023cross}, the French CS dataset~\cite{liu2018automatic}, and the British English CS dataset~\cite{sankar2022multistream}. The MCCSD contains three subsets: a six-cuer normal-hearing subset (6,000 samples), a single normal-hearing cuer subset (1,000 samples), and a single hearing-impaired cuer subset (818 samples). The French CS dataset contains 238 sentences from a single normal-hearing cuer, utilizing eight hand shapes and five positions to encode thirty-five phonemes. The British English CS dataset has both single-cuer (97 samples) and multi-cuer (390 samples) configurations. Its coding system maps forty-four phonemes through eight hand shapes and four positions.

However, prior research indicates that individuals with hearing impairments exhibit distinct lip movement patterns compared to normal cuers, posing enhanced challenges for ACSR systems~\cite{liu2023cross}. To evaluate methods in authentic communication contexts, we collect the new Multi-Hearing-Impaired Mandarin Chinese CS Dataset (MHI-MCCSD) as an extension. Aligning with MCCSD's collection protocol, eight native Mandarin cuers with hearing impairments (three females, five males) were recruited for data acquisition.

Recordings were conducted in soundproof environments across multiple weeks to capture natural lighting and participants' appearance variations. The linguistic content comprises 1,010 unique sentences spanning daily dialogues, literary excerpts (poems/idioms), and news transcripts, with word lengths varying from four to twenty-five words. As detailed in Table~\ref{table:dataset}, three cuers completed the full sentence set while the remaining five contributed partial recordings, resulting in a total of 5,272 annotated samples. This establishes MHI-MCCSD as the largest publicly available multi-cuer dataset representing hearing-impaired communication dynamics. Following MCCSD's partitioning strategy, data were randomly divided into training and test sets (4:1 ratio) without sentence overlap. All videos maintain 1280×720 resolution at 30 fps.

This new dataset enables a comprehensive evaluation of ACSR systems under realistic conditions while addressing critical gaps in representing hearing-impaired communication patterns. 

\begin{table*}[t] \small
\caption{Comparison with other methods on subsets of MCCSD~\protect\cite{liu2023cross} and MHI-MCCSD. The table reports the \textcolor{blue}{phoneme sequences}' CER (\%) and WER (\%) in different settings. The labels in brackets show the different cuer settings in subsets. The format refers to the number of cuers and whether they are hearing-impaired. ``HI'' or``H'' means cuers are hearing-impaired or not. Our Cued-Agent achieves good performance in all settings without training any fusion module.}
\centering

\tabcolsep 0.12 in
\begin{tabular}{c|cc|cc|cc|cc}

\toprule[1.3pt]
Dataset & \multicolumn{2}{|c|}{MCCSD (1-H)}& \multicolumn{2}{|c|}{MCCSD (1-HI)}& \multicolumn{2}{|c|}{MCCSD (6-H)} &\multicolumn{2}{|c}{MHI-MCCSD (8-HI)} \\
\midrule[1pt]
{\diagbox{Methods}{Metrics}} & CER $\downarrow$ & WER $\downarrow$ & CER $\downarrow$ & WER $\downarrow$ & CER $\downarrow$ & WER $\downarrow$ & CER $\downarrow$ & WER $\downarrow$ \\
\midrule[1pt]
ResNet18 + MHSA~\cite{vaswani2017attention} & 26.19 & 61.87 & 66.7 & 98.67 &61.83& 94.34& 82.03 & 99.91 \\
CMML~\cite{liu2023cross}& 9.81 & 25.54 & 32.23 & 69.45 & 30.01 &68.12  & 51.8& 91.26 \\
EcoCued~\cite{liu2024computation}& 9.54 & 25.03 &29.56 &61.59 & 29.75 & 67.83 &50.52 & 90.13\\
STF-ACSR~\cite{huang2025lendhandsemitrainingfree}& 
$\textbf{1.82}$&
$\textbf{5.19}$&
$\textbf{4.62}$&
$\textbf{12.21}$&
$\textbf{8.35}$&
$\underline{21.06}$&
$\textbf{10.96}$&
$\textbf{25.67}$\\
\midrule[1pt]

\textbf{Cued-Agent (Ours)} &$\underline{2.61}$&$\underline{6.56}$&$\underline{6.72}$&$\underline{16.23}$& $\underline{9.05}$&$\textbf{20.54}$&$\underline{12.67}$ &$\underline{29.86}$\\

\bottomrule[1.3pt]
\end{tabular}
\label{table:compare}
\end{table*}

\begin{table*}[t] \small
\caption{Performance on new proposed \textcolor{blue}{word sentence indicators} S-WER(\%) and Semantic Score (\%). 
Since Cued-Agent is the first method to realize the conversion from CS videos to word sentences, we hope it can serve as the baseline for future works.
}
\centering

\tabcolsep 0.015in
\begin{tabular}{c|cc|cc|cc|cc}
\toprule[1.3pt]
Dataset   & \multicolumn{2}{|c|}{MCCSD (1-H)}& \multicolumn{2}{|c|}{MCCSD (1-HI)}& \multicolumn{2}{|c|}{MCCSD (6-H)} &\multicolumn{2}{|c}{MHI-MCCSD (8-HI)} \\
\midrule[1pt]
{\diagbox{Methods}{Metrics}} & S-WER$\downarrow$& Semantic Score$\uparrow$&S-WER$\downarrow$&Semantic Score$\uparrow$&S-WER$\downarrow$&Semantic Score$\uparrow$&S-WER$\downarrow$&Semantic Score$\uparrow$\\
\midrule[1pt]
\textbf{Cued-Agent (Ours)} &12.1& 89.23 &24.38 & 76.48&31.28 &70.75 &40.84 &  60.25\\
\bottomrule[1.3pt]
\end{tabular}
\label{table:newidicator}
\end{table*}

\section{Experiments}
\label{sec:experiments}

\subsection{Experiment Setups}
\noindent \textbf{Datasets.}
Experiments are conducted on all MCCSD subsets and our proposed MHI-MCCSD. The details of the dataset settings can be found in Tab.~\ref{table:dataset}

\noindent \textbf{Implementation Details.} 
Two Python packages, retina-face~\cite{serengil2020lightface} and mediapipe2~\cite{lugaresi2019mediapipeframeworkbuildingperception}, are used to obtain the hand and lip ROI sequences from original CS videos. We use GPT-4o (2024-08-06)~\cite{gpt4o} as the MLLM for the Hand Recognition agent. The slow-motion threshold $\sigma$ is set to 6, and the index distance threshold $\theta$ is set to 2. To prevent post-processing challenges caused by disorganized outputs, we activated the structured output feature in the API. This ensures that GPT-4o predicts hand and position categories for keyframes as integer values and delivers the results in JSON format.
For the Lip Recognition agent and the Hand Prompt Decoding agent, we use the pretrained model from \cite{ma2023auto} 
as the base model. Specifically, in the Hand Prompt Decoding Agent, the hand prompt weight $\lambda_{prompt}$ is set to 4.5, and the tunable parameter $\lambda_{decode}$ is set to 0.5. The fine-tuning and decoding process are conducted on NVIDIA RTX A6000 GPUs. For the Self-Correction P2W Agent, we use the DeepSeek-R1~\cite{guo2025deepseek} to utilize its reasoning capabilities to achieve self-correction phoneme-to-word conversion.

\noindent \textbf{Evaluation Metrics.} We compare our Cued-Agents with several representative ACSR works, including the first Transformer-based method CMML~\cite{liu2023cross}, the computation-efficient EcoCued~\cite{liu2024computation}, and the most recent STF-ACSR~\cite{huang2025lendhandsemitrainingfree} with the SOTA performance. We also formed a Resnet18 with Multi-Head Self-Attention (MHSA)~\cite{vaswani2017attention} layers as the baseline. The Character Error Rate (CER), calculated as the ratio of edit distance to the total number of characters, and Word Error Rate (WER), similarly based on the word segmentation results, are used to evaluate the performance at the phoneme sequence level. In the newly proposed sentence-level evaluation, we use the pretrained Sentence-BERT~\cite{reimers-2019-sentence-bert} as the $\mathcal{M}(\cdot)$ to calculate the embeddings of predicted sentences and ground truth. 

\begin{table*}[t] \small

\caption{Ablation study on hand information and the Self-Correction P2W agent. The ``Dataset'' row shows the different subset cuer settings, where the format means the number of cuers and whether they are hearing-impaired. Cued-Agent effectively processed hand information to improve performance from the pure lip model significantly. The Self-Correction P2W agent effectively reduced the error rate of phoneme sequences by multiple rounds of revision with semantic information.}
\centering

\tabcolsep 0.09in
\begin{tabular}{c|cc|cc|cc|cc}
\toprule[1.3pt]
Dataset   & \multicolumn{2}{|c|}{MCCSD (1-H)}& \multicolumn{2}{|c|}{MCCSD (1-HI)}& \multicolumn{2}{|c|}{MCCSD (6-H)} &\multicolumn{2}{|c}{MHI-MCCSD (8-HI)} \\
\midrule[1pt]
{\diagbox{Methods}{Metrics}} & CER $\downarrow$ & WER $\downarrow$ & CER $\downarrow$ & WER $\downarrow$ & CER $\downarrow$ & WER $\downarrow$ & CER $\downarrow$ & WER $\downarrow$ \\
\midrule[1pt]
Pure Lip Information &3.64& 10.36 &10.18 & 25.61&12.81 &32.74 &18.41 & 43.79\\
+ Hand Information &2.97& 8.55 &8.28 &22.1& 9.5&24.74&13.86 & 33.97\\
+ Self-Correction P2W Agent & 2.61 & 6.56 &6.72&16.23& 9.05 & 20.54&12.67 & 29.86\\
\bottomrule[1.3pt]
\end{tabular}
\label{table:ablation}
\end{table*}

\subsection{Comparative Studies}
Tab. \ref{table:compare} shows the comparison results with other ACSR methods. Cued-Agent achieves comparable results with the current SOTA STF-ACSR and is much better than all Transformer-based methods. More precisely, Cued-Agent achieves performance comparable to STF-ACSR across all experimental settings. Notably, Cued-Agent requires only lip data for fine-tuning, and its Hand Prompt Decoding agent enables the first hand-lip joint decoding mechanism without 
requiring further training or additional parameters. This innovation significantly reduces the costs associated with hand recognition in training datasets compared to STF-ACSR. Cued-Agent even achieved the best WER value (20.54\%) in the MCCSD (6-H) experiment, confirming the effectiveness of using semantic information and CS system rules to correct phoneme sequences (i.e., the function of Self-Correction P2W agent). 

The best-performing Transformer-based method is Eco-Cued, but its performance is significantly lower than that of Cued-Agent. This proves that our Cued-Agent effectively avoids the performance degradation caused by data limitations via simplifying the complexity of hand recognition tasks and fusion modules.

\begin{figure*}[t]
\includegraphics[width=0.88\linewidth]{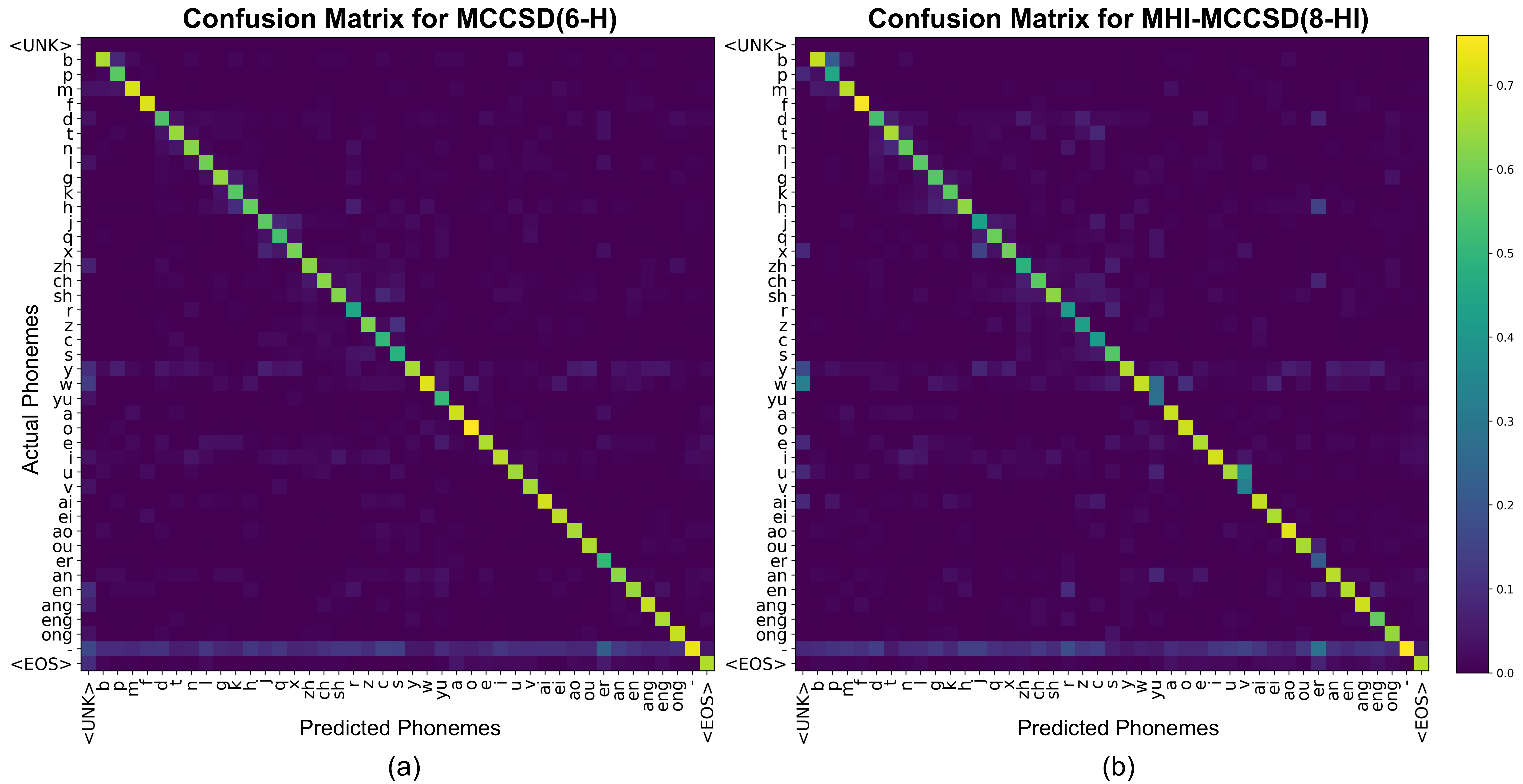}
\centering
\caption{Phoneme confusion matrices of Cued-Agent on the multi-cuer test sets of MCCSD~\protect\cite{liu2023cross} (Shown in (a)) and newly proposed MHI-MCCSD (Shown in (b)). Results reveal phonemes that are easily confused.}
\label{fig:confusion} 
\end{figure*}

\subsection{Performance on Sentence Metrics}
In Tab.~\ref{table:newidicator}, we first reveal the direct accuracy and semantic consistency of the ACSR model on sentence-level using the newly proposed metric. Cued-Agent is the first method that can convert CS visual input into natural sentences and is the only method in the table. The experimental results show the same trend as the results of phoneme-level evaluation. However, it cannot be ignored that S-WER has a significantly increasing value compared with phoneme-level WER, which shows that there is still a clear gap in converting phoneme sequences into language sentences.

\subsection{Ablation Studies}

\noindent \textbf{Effect of Hand Information.} In Cued-Agent, lip features are obtained by the Lip Recognition agent. If we directly decode the lip feature without hand information, we can get the performance with pure lip information as shown in the first row of Tab.~\ref{table:ablation}. When we try to add hand information, which corresponds to enabling the Hand Recognition agent and Hand Prompt Decoding agent in Cued-Agent, our method shows a significant decrease in error rate in all tests (shown in the second row of Tab.~\ref{table:ablation}). This proves that the system's hand-related agents can effectively obtain hand information and integrate it well with the lip feature.

Notably, in experimental settings involving hearing-impaired participants, hand movement information contributes significantly to enhanced performance. This observation aligns with the inherent challenges of accurately recognizing lip movements in hearing-impaired individuals and effectively demonstrates Cued-Agent's capability in processing hand movements. The system's enhanced performance when incorporating hand data underscores its effectiveness in capturing critical complementary information that lip reading alone might miss within this specific population.

\noindent \textbf{Effect of Self-Correction P2W Agent.} 
The experimental results in the second and third rows of Tab.~\ref{table:ablation} fully demonstrate that the design of the Self-Correction P2W Agent effectively corrects the phoneme sequence, resulting in improved performance in all test settings. The agent uses semantic and grammatical information to correct the phoneme sequence, effectively modifying the phonemes that do not conform to grammatical norms and semantic coherence, resulting in significantly improved word-level performance (WER).

\noindent \textbf{Discussion of Normal-Hearing and Hearing-Impaired.} 
To visually compare how normal-hearing data and hearing-impaired scenarios affect the ACSR model, we analyzed Cued-Agent's performance on the multi-speaker MCCSD subsets and MHI-MCCSD test sets using phoneme confusion matrices (Fig.~\ref{fig:confusion}). The matrices include 40 Chinese CS phonemes and symbols, with rows representing actual phonemes and columns indicating model predictions. Distinct off-diagonal color blocks highlight phonemes that the model frequently confused during recognition. 

Intuitively, Cued-Agent's performance in multi-hearing-impaired scenarios is slightly weaker than in normal-hearing scenarios. It is easier to confuse phonemes with the same hand codes, such as `b' and `p', `yu' and `w', and `v' and `u'.

\section{Conclusion}
\label{sec:conclusion}
This work introduces Cued-Agent, the first multi-agent system for ACSR, addressing the critical challenge of semantic ambiguity in hearing-impaired communication. By designing four sub-agents (Hand Recognition agent, Lip Recognition agent, Hand Prompt Decoding agent, and Self-Correction P2W agent), Cued-Agent achieves training-free multimodal alignment and CS phoneme-aware error correction. Experimental results demonstrate that Cued-Agent outperforms SOTA methods while eliminating costly training. Furthermore, the expanded Mandarin CS dataset, collected from hearing-impaired users, establishes a realistic benchmark for future assistive technology research. Future work will consider extending this method to the ACSR of other languages.

\clearpage
\bibliographystyle{ACM-Reference-Format}
\balance
\bibliography{egbib}

\end{document}